\documentclass{article}

\usepackage{arxiv}
\usepackage[utf8]{inputenc} 
\usepackage[T1]{fontenc}    
\usepackage[hidelinks]{hyperref}       
\usepackage{url}            
\usepackage{booktabs}       
\usepackage{amsfonts}       
\usepackage{nicefrac}       
\usepackage{microtype}      
\usepackage{lipsum}		
\usepackage{graphicx}
\usepackage{doi}
\usepackage{caption}
\usepackage{subcaption}
\usepackage{tikz}
\usetikzlibrary{shapes.geometric, positioning}
\graphicspath{ {./images/} }
\usepackage{siunitx}

\usepackage[shortcuts,acronym]{glossaries}
\newacronym{lasso}{LASSO}{least absolute shrinkage and selection operator}
\newacronym{nwp}{NWP}{numerical weather prediction}
\newacronym{ml}{ML}{machine learning}
\newacronym{svr}{SVR}{support vector regression}
\newacronym{svm}{SVM}{support vector machine}
\newacronym{mlp}{MLP}{multi-layer perceptron}
\newacronym{gbrt}{GBRT}{gradient boosting regression tree}
\newacronym{mae}{MAE}{mean absolute error}
\newacronym{rmse}{RMSE}{root mean squared error}
\newacronym{nrmse}{nRMSE}{normalized root mean squared error}
\newacronym{mse}{MSE}{mean squared error}
\newacronym{kld}{KLD}{Kullback-Leibler Divergence}
\newacronym{pv}{PV}{photovoltaic}
\newacronym{wrt}{w.r.t.}{with respect to}
\newacronym{pdf}{PDF}{probability density function}
\newacronym{mtl}{MTL}{multi-task learning}
\newacronym{stl}{STL}{single-task learning}
\newacronym{sps}{SPS}{soft parameter sharing}
\newacronym{hps}{HPS}{hard parameter sharing}
\newacronym{csn}{CSN}{cross-stitch network}
\newacronym{sn}{SN}{sluice network}
\newacronym{ern}{ERN}{emerging relation network}
\newacronym{nlp}{NLP}{natural language processing}
\newacronym{lstm}{LSTM}{long-tem short memory}
\newacronym{tcn}{TCN}{temporal convolution network}
\newacronym{cnn}{CNN}{convolutional neural network}
\newacronym{tl}{TL}{transfer learning}
\newacronym{dtw}{DTW}{dynamic time warping}
\newacronym{elbo}{ELBO}{expected lower bound}
\newacronym{csge}{CSGE}{coopetitive soft gating ensemble}
\newacronym{ae}{AE}{autoencoder}
\newacronym{cnnae}{CAE}{convolutional autoencoder}
\newacronym{vae}{VAE}{variational autoencoder}
\newacronym{dae}{DAE}{denoising autoencoder}
\newacronym{pca}{PCA}{principal component analysis}
\newacronym{rl}{RL}{representation learning}
\newacronym{sgd}{SGD}{stochastic gradient decent}
\newacronym{ttl}{TTL}{transductive transfer learning}
\newacronym{itl}{ITL}{inductive transfer learning}
\newacronym{sqtl}{SQTL}{sequential transfer learning}
\newacronym{aic}{AIC}{Akaike information criterion}
\newacronym{bic}{BIC}{Bayesian information criterion}
\newacronym{bma}{BMA}{Bayesian model averaging}
\newacronym{elm}{ELM}{exterem learning machine}
\newacronym{belm}{BELM}{Bayesian exterem learning machine}
\newacronym{relu}{ReLU}{Rectified Linear Unit}
\newacronym{DBN}{DBN}{deep belief network}
\newacronym{RBM}{RBM}{restricted Boltzmann machines}
\newacronym{RBF}{RBF}{radial-basis function}
\newacronym{blr}{BLR}{Bayesian linear regression}
\newacronym{sp}{SP}{source parameters}
\newacronym{crps}{CRPS}{continuous ranked probability score}
\newacronym{ghi}{GHI}{global horizontal irradiance}
\newacronym{dni}{DNI}{direct normal irradiance}
\newacronym{dhi}{DHI}{diffuse horizontal irradiance}
\newacronym{iconeu}{ICON-EU}{Icosahedral Nonhydrostatic-European Union}
\newacronym{synop}{SYNOP}{surface synoptic observation}
\title{Synthetic Photovoltaic and Wind Power Forecasting Data}

\author{ \href{https://orcid.org/0000-0002-3230-8822}{\includegraphics[scale=0.06]{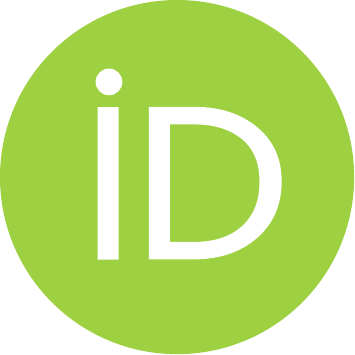}\hspace{1mm}Stephan~Vogt}
\\
	Intelligent Embedded System Lab\\
	Kassel University, Germany\\
	Email:  \texttt{stephan.vogt@uni-kassel.de} \\
	\And
	\href{https://orcid.org/0000-0002-9979-8053}{\includegraphics[scale=0.06]{orcid.pdf}\hspace{1mm}Jens~Schreiber} \\
	Intelligent Embedded System Lab\\
	Kassel University, Germany\\
	Email:  \texttt{j.schreiber@uni-kassel.de} \\
	\And
	\href{https://orcid.org/0000-0001-9467-656X}{\includegraphics[scale=0.06]{orcid.pdf}\hspace{1mm}Bernhard~Sick} \\
	Intelligent Embedded System Lab\\
	Kassel University, Germany\\
	Email:  \texttt{b.sick@uni-kassel.de} \\
}

\renewcommand{\shorttitle}{Technical Report}

\hypersetup{
pdftitle={A template for the arxiv style},
pdfsubject={q-bio.NC, q-bio.QM},
pdfauthor={Stephan Vogt},
pdfkeywords={Synthetic Data, Wind / Photovoltaic Energy, Forecasting},
}

\tikzstyle{decision} = [diamond, draw, fill=red!15, text width=4.5em, text badly centered, node distance=15ex, inner sep=0pt]
\tikzstyle{block} = [rectangle, draw, fill=blue!70!red!10, node distance=15ex, text centered, rounded corners, text width=25ex, minimum height=5ex]
\tikzstyle{line} = [draw, -latex]
\tikzstyle{cloud} = [draw, rectangle,fill=green!20, node distance=25ex, minimum height=5ex, rounded corners]

\begin{document}
\maketitle

\begin{abstract}
    Photovoltaic and wind power forecasts in power systems with a high share of renewable energy are essential in several applications.
    These include stable grid operation, profitable power trading, and forward-looking system planning.
    However, there is a lack of publicly available datasets for research on \acl{ml}-based prediction methods.
    This paper provides an openly accessible time series dataset with realistic synthetic power data.
    Other publicly and non-publicly available datasets often lack precise geographic coordinates, timestamps, or static power plant information, e.g., to protect business secrets.
    On the opposite, this dataset provides synthetic time-series measurements, input features, geographic information, timestamps, and physical properties of a set of individual power plants.
    %
    %
    The dataset comprises 120 photovoltaic and 273 wind power plants with distinct sides all over Germany from 500 days in hourly resolution.
    This large number of available sides allows forecasting experiments to include spatial correlations and run experiments in transfer and multi-task learning.
    Furthermore, the dataset includes side-specific, power source-dependent, non-synthetic input features from the \acl{iconeu}.
    However, since timestamps and location are known, combining the synthetic power measurements with other weather models is also possible.
    A simulation of virtual power plants with physical models and actual weather measurements provides synthetic power measurement time series.
    These time series correspond to the power output of virtual power plants at the location of the respective weather measurements.
    Since the synthetic time series are based exclusively on weather measurements, possible errors in the weather forecast are comparable to those in actual power data.
    %
    %
    %
    %
    In addition to the data description, we evaluate the quality of weather-prediction-based power forecasts by comparing simplified physical models and a \acl{ml} model.
    This experiment shows that forecasts errors on the synthetic power data are comparable to real-world historical power measurements.
    The \acl{ml} model consists of gradient boosted regression trees, which show a lower forecast error for wind and photovoltaic with sufficient training data.
    In addition, we consider truncated datasets and find that, at least for photovoltaics, there is an advantage of the physical approaches when we only use a fraction of the training data.
    The errors serve as a baseline for future experiments.
\end{abstract}


\section{Introduction}
Renewable energies are taking on an increasing role in modern energy supply systems.
As their share increases, the energy supply system's volatility and dependence on the weather grow.
Therefore, energy forecasts are an essential tool to assure grid stability and reduce uncertainty, e. g., in energy markets or planning assets of electrical grids.
For this reason, the importance of research on forecasts of wind energy and \ac{pv} systems has increased.

In renewable energy forecasts, we typically utilize weather forecasts, such as the wind speed or radiation, from a so called~\ac{nwp} model.
These predicted weather features are the input to \ac{ml} models, predicting the expected power generation, e. g., in \textit{day-ahead} forecasts between $24$ and $48\si{\hour}$ into the future.
Datasets must be open for researchers from different fields to develop new forecasting techniques that reduce forecast error and decrease uncertainty in the energy system.
Open datasets allow to compare newly developed methods, reproduce these results, and compare results to state of the art.
\newline
However, the existing datasets usually have various limitations.
In some cases, the number of sites considered is relatively small (e.g., in \cite{hong2014global}), so it is impossible to make statements with statistical significance.
In some cases, the data is anonymized without an exact geographic location and not synchronized in time, see \cite{EuropeWindFarm} and \cite{GermanSolarFarm}.
Such anonymizations prohibit a model from taking spatiotemporal dependencies into account that are common in the electrical grid.
Further static information, e.g., technical conditions such as the orientation of \ac{pv} modules or the rotor-generator ratio of wind turbines, is often missing.
At the same time, when a park has insufficient data for traditional ~\ac{ml} techniques, the utilized physical models require this static information.
Additionaly, these static information are also essential for, e.g., \ac{tl} and zero-shot learning techniques~\cite{Schreiber2021} that are~\ac{ml} methods to handle limited data.
In the future, more and more forecasting techniques only require limited data and provide high-quality forecasts to compensate for the severe increase in volatile energy resources.
Additionally, in practical application, i.e., in typical systems of forecast vendors, one has to deal with missing technical information.
On the other hand, in research, it is critical to have a dataset as complete as possible and thus to be able to analyze how the lack of such information potentially affects the forecast quality.
Therefore, we provide a complete and open\footnote{Currently, the dataset is not accessible through a central data-repository. Therefore, please write us an e-mail in case you require the data.} accessible dataset that allows researchers to answer various questions in the field of wind and photovoltaic forecasting with \ac{ml} techniques.
%

%
In particular, we provide a large dataset that comprises synthetic but realistic 120 photovoltaic and 273 wind virtual power plant sides for roughly 500 days that includes the most relevant static data.
This dataset is the largest open accessible dataset for renewable power forecasts to the best of our knowledge.
We assure that the synthetic power measurements are realistic by utilizing meteorological weather measurements and physical models.
This consideration is essential as most of the forecast error in energy forecasting relates to the forecast error caused by the input features from the~\ac{nwp} model.
The period under consideration starts December 8, 2018, and ends June 2, 2020 with hourly time resolution.
The length of 500 days allows a model training to include all seasons in the training set (a year comprises about 75\%) and still have sufficient test data (about 25\%).

The remainder of this article is structured as follows.
First, in~Sec.~\ref{sec_intended_use}, we overview the data and the intended use.
We also argue why there is a necessity for such a dataset, e.g., in the context of~\ac{tl}.
Afterward, we outline the most relevant types of data utilized for the generation of the dataset in Sec.~\ref{sec_data_basis}.
Based on those definitions, we detail the approaches to generate synthetic power measurements in Sec.~\ref{sec_synthetic_data_generation}.
Sec.~\ref{sec_experimental_results} shows our results for power forecasting techniques on the dataset as a baseline for future research.
Finally, we summarize our work and suggest future work in Sec.~\ref{sec_conclusion_future_work}.

\section {Data Overview and Intended Use}\label{sec_intended_use}
The data per location is divided into input data (\verb|data_input_<location id>.csv|) and target data (\verb|data_target_<location id>.csv|).
Both CSV files, input, and target have a timestamp that allows a unique assignment to each other.
This timestamp is the prediction time, i.e., the time the prediction is valid.
The simplest option is to treat the data as i.i.d. samples (even though they are not).
This assumption makes it possible to treat the problem as a non-linear but time-independent regression problem, and many standard supervised \ac{ml} methods can be applied.
Here, the input CSV represents a design matrix (input data matrix) whose individual rows correspond to the rows of the target CSV and the columns correspond to suitable input features.
One might not use the raw timestamp as an input feature.
Otherwise, the model possibly never generalizes beyond the training period (i.e., the test period, for example).
The target-CSV includes a column 'pw', which contains the synthetic power measurement and thus the intended target.
We generate these targets through a physical model based on \textit{real-world} on-site meteorological weather measurements.
As most of the forecast errors in energy forecasts relate to the~\ac{nwp} forecasts, by utilizing real-world meteorological weather measurements, we are capable of providing realistic synthetic power measurements.
\newline
Furthermore, we provide a test flag and the output of a physical baseline model.
To communicate a forecasting model's final test errors, we recommend the data samples with the test flag.
We utilize the same physical model and parametrization that generated the synthetic power for the baseline.
However, these forecast \ac{iconeu}~\ac{nwp} forecasts instead of the real-world weather measurements.
\newline
Beginning December 1, 2019, a flag marks the beginning of a predefined test period.
The test flag should help to better compare results across research groups.
We suggest using samples between December 8, 2018, and December 1, 2019, for training and validation (model selection).
The test data from a period after December 1, 2019, enables the test of a hypothesis under consideration as independently as possible of the training period.
However, it is noteworthy that since the test data were published together with the training data, indirect overfitting may occur unintentionally, for example, if the test dataset is taken into account in some way in the model selection.
Therefore, methods compared in this way might not work with a similar advantage on future, previously unknown data.
Nevertheless, it can help to assess the capabilities of investigated methods better.

\subsection{Forecasting Scenarios}
We created the dataset with a day-ahead forecasting scenario in mind, which is typical for many forecasting vendors.
At the same time, day-ahead forecasts are more challenging than intra-day forecasts, as with an increasing forecast horizon of the weather model, the error of the input features increases.
%
%
Based on current literature~\cite{Alkhayat2021} and exchanges with experts from the industries, we identified the following key research areas:

\begin{itemize}
    \item \textbf{Single task forecasting} is still ongoing research and recent developments from the field of deep learning allow to reduce the forecast error and thereby decrease the uncertainty for the electrical grid.
    \item Regional estimation of wind / \ac{pv} power production, which take \textbf{geospatial dependencies} into account, better capture the dynamics in the electrical grid.
    \item \textbf{Multi-task learning} reduces the forecast error, the number of parameters, and the training time~\cite{Schreiber2020,Vogt2019}. Effectively, this technique helps in compensating for the high demand for forecasts.
    \item \textbf{Transfer learning} allows for reducing the training time and the required number of historical power measurements that are often not available~\cite{Vogt2019,Schreiber2021}.
    \item \textbf{Meta-learning} and similarly \textbf{zero-shot learning} ideally allow to provide power forecasts without any historical data~\cite{Schreiber2021} and are, e.g., an essential technique to provide forecasts for new power plants without any historical data.
    \item \textbf{Explainable}~\ac{ml} allows participants in the energy systems to make justified decisions.
\end{itemize}
\section{Data Basis}\label{sec_data_basis}
Three different data sources play a role in synthesizing power forecasting data, namely meteorological measurements, numerical weather predictions, and static (metadata) data for the individual power plants.
The static data forms the parametrization of the physical models.
Furthermore, this model simulates the power output based on meteorological measurements.
Besides, the numerical weather predictions provide the necessary input features, which currently form our best source for information about future weather conditions in a short-term range (within a few days).

\subsection{Meteorological Measurements}
\begin{figure}[ht]
    \centering
    \includegraphics[width=0.4\textwidth]{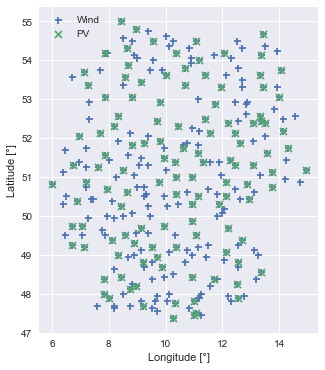}
    \caption{Geographical locations of weather measurements, respective sides of the synthetic power measurements.}
    \label{fig:geo_locations}
\end{figure}

Since photovoltaic and wind energy depends highly on the local weather conditions, meteorological data form the basis for the synthetic data.
The power output of a single plant forms the target variable in many photovoltaic and wind power forecasting tasks.
Typically, power plants have a quick response to changes in weather conditions, and therefore the variability of the relevant weather variables is almost directly transferred to the power output.
Meteorological measurements, such as wind speed or global horizontal irradiation, can provide a realistic degree of variability and randomness as the basis for the target data.
\newline
In addition, meteorological measurements accurately represent spatial and temporal relationships of the underlying atmospheric state in equal measure.
Figure \ref{fig:geo_locations} shows the spatial distribution of the measurement sites.
At the same time, these sides represent the virtual locations of the power plants, i.e., wind energy or photovoltaic power plants, which correspond to the synthetic data.
The plot shows that wind speed measurements are also available at all locations where radiation data is available.
The reverse is not applicable, so less than half of the sites have radiation measurements.
The underlying measurements for solar are from \cite{synopSolar2020} and the wind resource measurements from~\cite{synopWind2020}.
\subsection{Input Features and Numerical Weather Prediction}
We take input features from the \ac{iconeu} weather model as a basis, described in \cite{reinert2016icon}.
These numerical weather predictions were computed and provided by the German Meteorological Service (\emph{Deutscher Wetterdienst}) and provided for this study by Fraunhofer IEE.
For \ac{pv}- and wind energy, several features were extracted and processed individually.
In contrast to the \ac{synop} measurements, the features from the \ac{iconeu} model are available within the considered forecast horizons (day-ahead scenario) with hourly resolution.
The hourly resolution available here forms the basis for the final merged dataset.
\subsubsection*{Photovoltaic Input Feature}
The features for the \ac{pv} forecast are listed below, with feature names based on those in \cite{reinert2016icon}:
\begin{description}
    \item[fcst\_time] Forecast date and time for which the forecast is to be generated and to which the below-mentioned features refer to in the format \verb|yyyy-mm-dd HH:MM:SS| [UTC].
    \item[nwp\_fcst\_horiz\_hours] forecast horizon in relation to the start of the numerical weather prediction model run at 00:00 of the previous day [h].
    \item[T\_HAG\_2\_M] Air temperature 2m above the earth's surface [°K].
    \item[RELHUM\_HAG\_2\_M] Relative humidity 2m above earth surface [\%].
    \item[PS\_SFC\_0\_M] Air pressure at the earth's surface [Pa].
    \item[U\_GVL\_60\_HL] Zonal wind speed (wind parallel to latitude) at about 10m above ground [m/s].
    \item[V\_GVL\_60\_HL] Meridional wind speed (longitude-parallel wind) at about 10m above ground [m/s].
    \item[ASWDIFDS\_SFC\_0\_M\_INSTANT] Instantaneous solar diffuse radiation at the earth's surface with a time lag relative to the forecast time of -1h, 0h, or +1h (three features with the suffixes \_m1,\_, or \_p1) [W/$m^2$].
    \item[ASWDIRS\_SFC\_0\_M\_INSTANT] Instantaneous solar direct radiation at the earth's surface with a time lag related to the prediction time by -1h, 0h or +1h (three features with suffixes \_m1,\_ or \_p1) [W/$m^2$].
    \item[solar\_azimuth] Azimuth angle of the sun position calculated using prediction time and the toolbox PVLib, described in \cite{Holmgren2018} [°].
    \item[solar\_zenith] Zenith angle of the sun position calculated using prediction time and Toolbox PVLib.
\end{description}
Except for the prediction time \verb|fcst_time|, these features can be used directly as inputs for nonlinear regression models.
We use \verb|fcst_time| to assign the time to the corresponding target value, to other weather models, or to calculate additional features, e.g., representing the season.
Nevertheless, \verb|fcst_time| is less suitable as a direct input feature in the case of \ac{ml} methods since the test data will differ in the considered time and therefore, an \ac{ml} model will most likely not generalize well.
\newline
Irradiance features ending with \verb|_INSTANT| differ from the original weather model quantities due to our post-processing of original irradiance values.
In the \ac{iconeu} weather model, these quantities are stored as mean values since the beginning of the model run.
Since we aim to predict the instantaneous power of the system, we convert these quantities to instantaneous irradiance features through the differences between temporally adjacent time steps.
\newline
In total, the synthetic \ac{pv} dataset thus contains 15 different features, 14 of which can be used directly as input to a regression model.
\subsubsection*{Wind Power Input Feature}
In the case of the synthetic wind power data, the data are available with 19 comparable input features, which we describe in detail in the following list:
\begin{description}
    \item[fcst\_time] Forecast date and time for which the forecast is to be generated and to which the below-mentioned features refer to in the format \verb|yyyy-mm-dd HH:MM:SS| [UTC].
    \item[nwp\_fcst\_horiz\_hours] is the forecast horizon in relation to the start of the numerical weather prediction model run at 00:00 of the previous day [h].
    \item[T\_HAG\_2\_M] Air temperature 2m above the earth's surface [°K].
    \item[RELHUM\_HAG\_2\_M] Relative humidity 2m above earth surface [\%].
    \item[PS\_SFC\_0\_M] Air pressure at the earth's surface [Pa].
    \item[U\_GVL\_58\_HL] Zonal wind speed (wind parallel to latitude) at about 100m above ground with a time lag related to the forecast time of -1h, 0h or +1h (three features with suffixes \_m1,\_ or \_p1) [m/s].
    \item[V\_GVL\_58\_HL] Meridional wind speed (longitude-parallel wind) at about 100m above ground with a time lag related to the forecast time of -1h, 0h or +1h (three features with suffixes \_m1,\_ or \_p1) [m/s].
    \item[U\_GVL\_60\_HL] Zonal wind speed (wind parallel to latitude) at about 10m above ground with a time lag related to the forecast time of -1h, 0h or +1h (three features with suffixes \_m1,\_ or \_p1) [m/s].
    \item[V\_GVL\_60\_HL] Meridional wind speed (longitude-parallel wind) at about 10m above ground with a time lag related to the forecast time of -1h, 0h or +1h (three features with suffixes \_m1,\_ or \_p1) [m/s].
    \item[ASWDIFDS\_SFC\_0\_M\_INSTANT] Instantaneous solar diffuse radiation at the Earth's surface [W/$m^2$].
    \item[ASWDIRS\_SFC\_0\_M\_INSTANT] Instantaneous solar direct radiation at the Earth's surface [W/$m^2$].
\end{description}
\subsection{Static Data}\label{sec:static_data}
We summarize the static data (or metadata) for all locations of the respective energy source in the file \verb|meta.csv|.
The individual rows in the CSV represent the static data for the individual locations.
For both wind and \ac{pv}, the file contains the identification number (\verb|loc_id|), the geographical coordinates (\verb|longitude| or \verb|long| and \verb|latitude| or \verb|lat|), the file names of the associated input and target data (\verb|input_file_name| and \verb|target_file_name|), and a brief statistic in form of the number of samples per sample set (\verb|num_train_samples|, \verb|num_test_samples|).
Further static data differ depending on the energy source, especially the physical parametrization.
\subsubsection*{Photovoltaic Parametrization}
In the case of photovoltaics, one finds information about the module, the inverter, and the wiring (modules per string and strings per inverter).
However, we selected only one configuration for simplicity, namely module: Canadian Solar CS5P 220M (2009), inverter: ABB Micro 0 25 I OUTD US 208 (208V), with one single module per string, and one string per inverter.
We chose this configuration based on the experience that the exact module and inverter parameters (e.g., rated power, efficiency, and temperature coefficients) have a relatively small impact on the prediction quality, especially after normalization and compared to the errors of the weather model.
Much more important, however, is the exact orientation of the plant.
The orientation parameters are drawn from a random distribution to have a diverse collection of different orientations in the dataset.
We sample the module tilt angle from a uniform distribution in the interval between 0° (flat) and 90° (vertical).
Similarly, we sample the azimuth angle from a uniform distribution between 90° (east) and 270° (west), with the south defined as 180°.
\subsubsection*{Wind Power Parametrization}
We summarize similar information about the simulated wind turbines in the corresponding \verb|meta.csv| file associated with the synthetic wind power dataset.
The file includes the hub height used for vertical interpolation, the rotor diameter in meters, and the nominal power in kilowatts.
Additionally, we provide information about the respective turbine type.
The latter information is crucial to be able to reproduce the dataset since the corresponding power curves were stored as manufacturer information in the simulation, similar to \cite{enercon2015}.
The turbine power curve and the hub height of each entry in \verb|meta.csv| were individually selected by randomly choosing one of the possible hub heights (either maximum or minimum) and one specific turbine type.

\section{Synthetic Data Generation}\label{sec_synthetic_data_generation}
\begin{figure}[ht]
    \centering
    \includegraphics{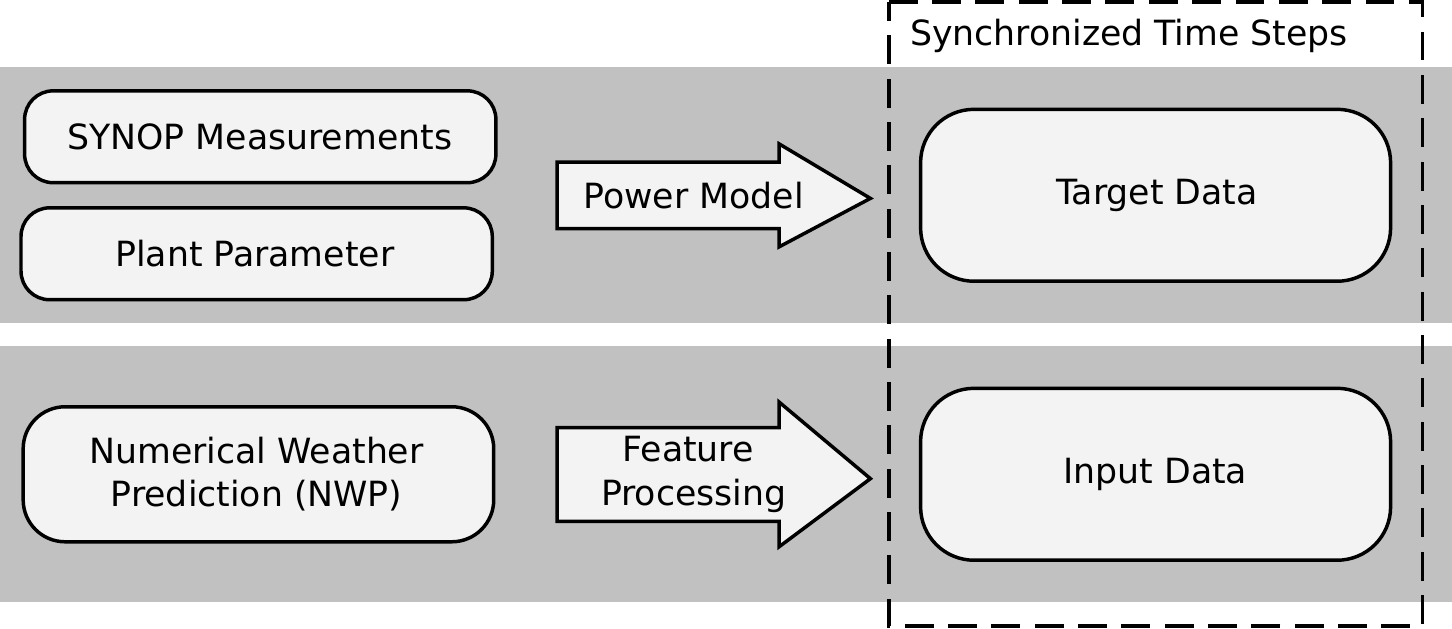}
    \caption{Flowchart for synthesing power data from weather measurements and of the feature extraction process.}
    \label{fig:flowchart}
\end{figure}

This section describes in detail how we assemble the datasets.
Fig.~\ref{fig:flowchart} depicts the overall concept.
On the left, various data sources are listed.
We process those data sources either through a physical model or feature pre-processing.
The resulting output of the synthetic data generation, namely target and input data, is given on the right.
\newline
The top row, the top grey box, shows the process to synthesize the target data.
In this process, a power model transforms meteorological measurements (\ac{synop}) and static plant configurations into a power time series.
The used power models are implemented either as a physical model (\ac{pv}) or are given as an empirical look-up table by the power plant manufacturers (power curves of wind turbines).
The resulting time series represents the synthetic power measurement, which we provide as a target for a \ac{ml}-based forecasting model.
Note that we do not use numerical weather prediction data to prepare the target data.
This process ensures that the expected forecast errors are similar in magnitude and characteristics to real-world power forecasts.
\newline
The bottom row, the bottom grey box, shows the process of input data preparation.
The raw input features include numerical weather predictions and the corresponding time stamp information at the nearest weather model grid node to the measurements side.
Depending on the energy source, certain features are selected and pre-processed.
%
%
%
%
This feature processing consists, for example, of the time-lags of specific input signals concerning the prediction time.
The motivation for this is to provide samples that can even be used in regression models that make an i.i.d. assumption.
Other features, such as sun angles in the case of photovoltaics, are also calculated and provided.
\newline
Furthermore, on the right side, there is a dashed box.
This box symbolizes the synchronization of the input data and the target variable.
This synchronization takes place according to a day-ahead forecast.
Here, a scenario was set, which assumes the daily creation of a day-ahead forecast.
More precisely, we assume that the 00:00 o'clock model run of the weather model is delivered to the power forecast provider.
Related to the origin of the 00:00 model run, we synchronize the forecast horizons between 24h and 47h inclusive with the power values with hourly resolution.
The hourly resolution of the weather prediction is adopted by down-sampling the weather measurements, which initially have a 10min time resolution.
Therefore, interpolation is not needed.
Furthermore, for each geographic power plant location, the predictions of the nearest point of the \ac{nwp} grid were extracted.

\subsection{Photovoltaic Power Model}
\begin{figure}[tb]
    \centering
    \includegraphics[width=0.98\textwidth]{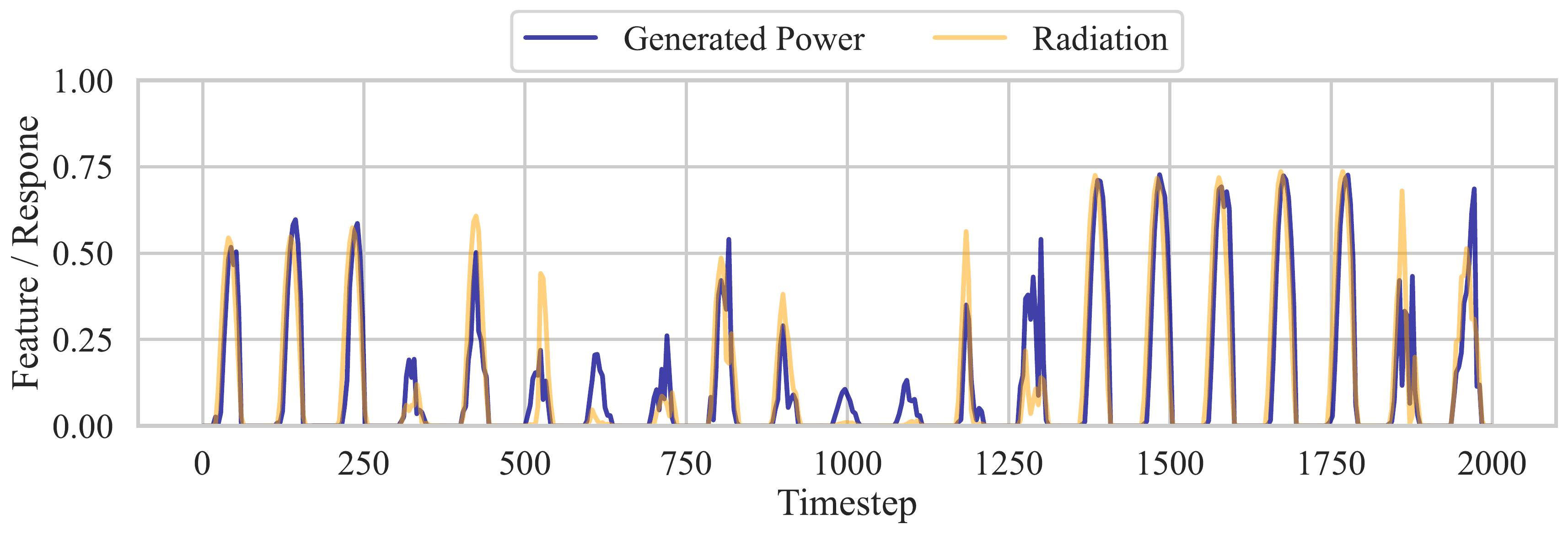}
    \caption{Exemplary three weekly course of the synthetic power generation of a virtual \ac{pv} power plant and the irradiation normalized with the nominal power and 1000 W/m². The data was linearly interpolated to a 15-minute resolution.}
    \label{fig:PVSYN_timeseries}
\end{figure}

Radiance measurements are converted via a physical model into power measurements to generate the synthetic power measurements.
The underlying physical model is the one from the simulation toolbox PVLib~\cite{Holmgren2018}.
All simulations use the same photovoltaic module type \emph{Canadian Solar CS5P-220M (220W) Solar Panel} which is specified in more detail in \cite{solarDesignTool2022}.
Similarly, the same type of inverter was used exclusively for all simulations, namely the \emph{MICRO-0.25/0.3/0.3HV-I-OUTD-US-208/240} described in \cite{inverter2014}.
We assume that all \ac{pv} power plants contain only a single module as an additional simplification.
The synthetic measurement time series are normalized by the installed capacity, i.e., the module peak power.
%
%
%
%
%
\newline
The \ac{synop}-power measurements include the ten-minute sum of incoming solar radiation in Joule per cubic Centimeter and the ten-minute sum of the diffuse solar radiation with the same unit.
PVLib expects the \ac{ghi}, the \ac{dni} and the \ac{dhi} in Watt per square Meter, as well as wind speed and air temperature.
To overcome this difference, we transformed the measured quantities accordingly by using the methods given in PVLib, i.e., \emph{pvlib.irradiance.dni} to estimate \ac{dni} based on \ac{ghi}, \ac{dhi} and Zenith angle.
Previously, the zenith angle was computed based on the geographic position and time, while \ac{ghi} and \ac{dhi} are transformed under the assumption of a discrete approximation of the energy-time derivative, i.e., that the power is the ratio of a discrete change in energy in a given time interval.
Wind speed and air temperature remained with their PVLib-default values of 0m/s and 20° Celsius for further simplification.
\newline
Finally, the simulation requires information about the location.
This information includes the geographic coordinates, which correspond to the coordinates of the weather measurement and the altitude.
For simplicity, we assume the altitude to be 0m above sea level.
All other model inputs, such as the albedo, were left at the PVLib default values.
\newline
Figure~\ref{fig:PVSYN_scatter} and Figure~\ref{fig:PVSYN_timeseries} show the simulated \ac{pv} power in relation to the predicted direct irradiation on a horizontal plane.
An increase in irradiation leads to an expected increase in the power generation in figure~\ref{fig:PVSYN_scatter}.
At the same time, the inherent uncertainty of the weather prediction and the missing consideration of the module orientation leads to a spread of the point cloud in the y-direction.
This difference is also a typical behavior in real-world datasets.
\newline
Figure~\ref{fig:PVSYN_timeseries} shows the normalized power output (blue line) of the same plant in a three-week window together with the normalized horizontal direct irradiation (yellow line).
The slight shift of the blue curve, compared to the yellow curve, to the right results from a slight west-ward orientation of the simulated \ac{pv} module.
While sunny days predominantly show agreement in the yellow and blue curve profile, cloudy days show more extensive and random deviations between the synthetic power data and the predicted solar resource.
That is typical and results from the \ac{nwp} difficulties to predict exact cloud positions.
Furthermore, the predicted solar resources appear to fluctuate less than the synthetic power generation.
This difference in fluctuation also results from the properties of the \ac{nwp}, which rather forecasts an expected value of a local patch (several thousand meters in size) compared to the local point of the single plant and, with that, to smooth out any detailed cloud structures.
\subsection{Wind Energy Power Model}
\begin{figure}
    \centering
    \begin{subfigure}[b]{0.48\textwidth}
        \centering
        \includegraphics[width=0.98\textwidth]{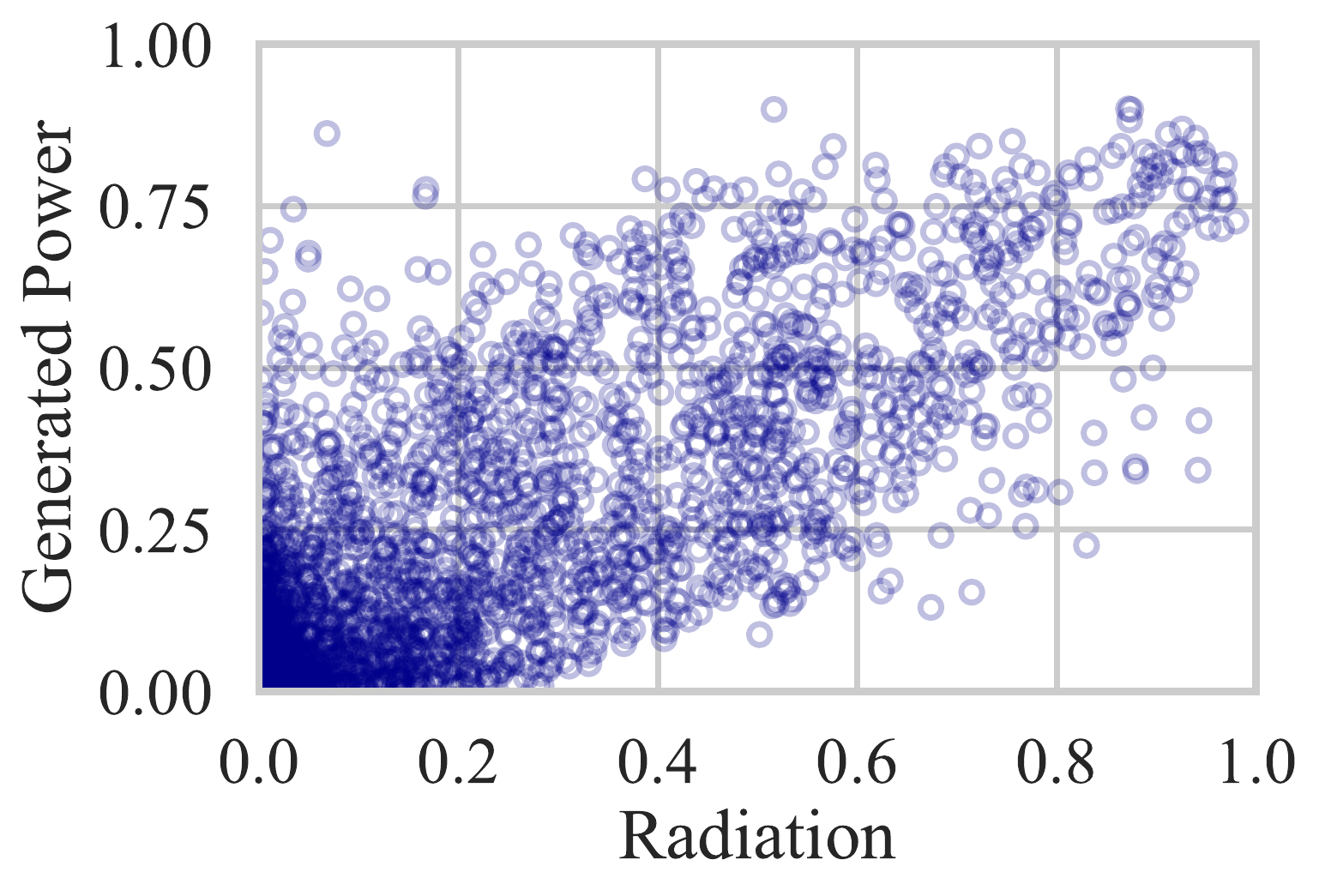}
        \caption{Feed-in of a virtual \ac{pv} plant plotted against direct-horizontal radiation (\ac{iconeu}).}
        \label{fig:PVSYN_scatter}
    \end{subfigure}
    \hfill
    \begin{subfigure}[b]{0.48\textwidth}
        \centering
        \includegraphics[width=0.98\textwidth]{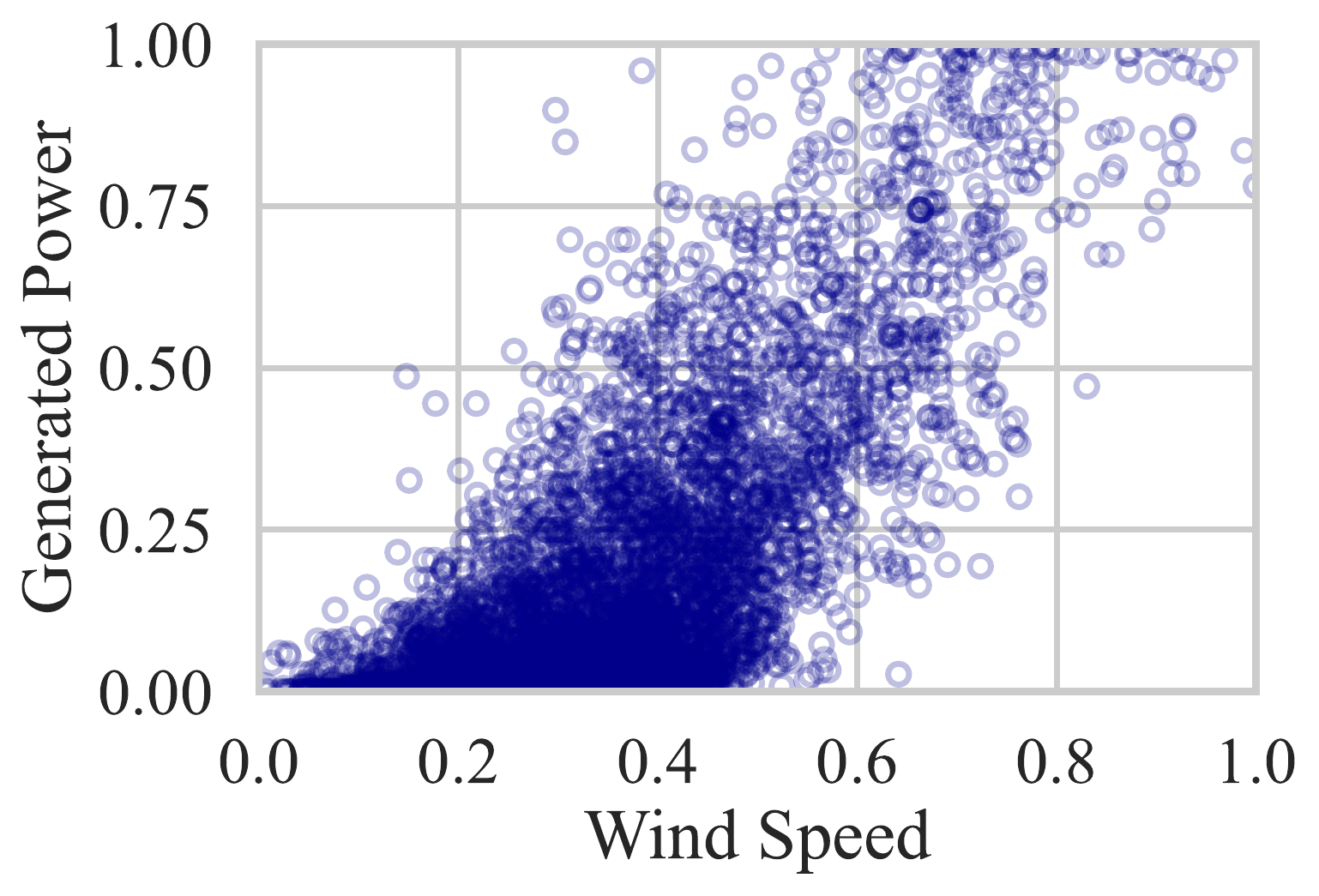}
        \caption{The output of a virtual wind turbine over the magnitude of the predicted wind speed vector (\ac{iconeu}).}
        \label{fig:WINDSYN_scatter}
    \end{subfigure}
    \caption{Scatter plots of the synthetic power generation (\ac{synop} based) over the corresponding predicted weather resource.}
\end{figure}

\begin{figure}[tb]
    \centering
    \includegraphics[width=0.98\textwidth]{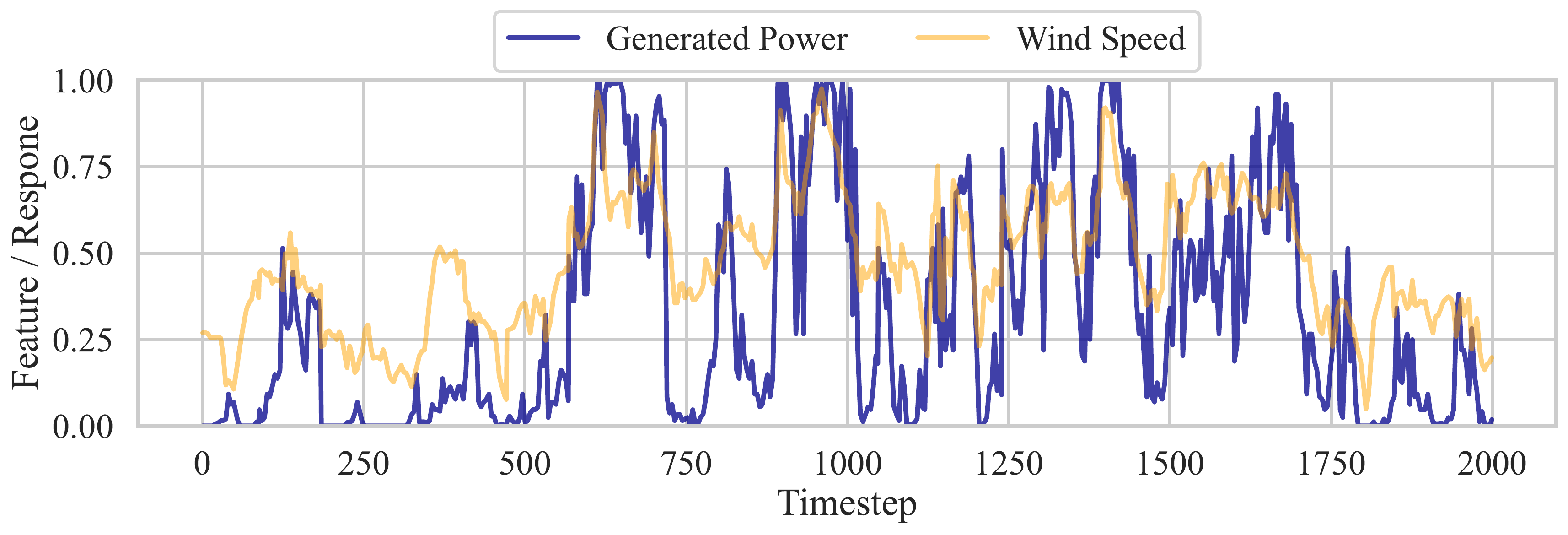}
    \caption{Three-week history of synthetic power generation from a virtual wind turbine and the predicted underlying wind speed normalized by the rated power and maximum wind speed, respectively. The data was linearly interpolated to a 15-minute resolution.}
    \label{fig:WINDSYN_timeseries}
\end{figure}

A simple power curve model was used to calculate wind power.
In the first step, we extrapolate the measured wind speed $u_{ref}$ from 10m height to the height of the turbine.
For this purpose, the so-called wind profile power law is applied, described in \cite{touma1977dependence}.
With that, we compute the wind speed at the height of the wind turbine $u_{ref}$ with
\begin{eqnarray}
    u = u_{ref}\left(\frac{z}{z_{ref}}\right)^{\alpha},
\end{eqnarray}
where the hub height $z$ of the wind turbine, the measurement height $z_{ref}=10m$, the measured wind speed $u_{ref}$ and a stratification coefficient $\alpha = \frac{1}{7}$ (assumption of neutral stratification) are used.
In the next step, a linear interpolation is applied together with a lookup table of the manufacturer's power curve as given in \cite{enercon2015}.
Each power curve lookup table is defined in terms of reference wind speeds and reference power measurements, discretized with a resolution of 1 m/s up to a maximum reference wind speed of 25 m/s.
Other possible impact factors, such as air density, turbulence, or atmospheric stability are ignored for simplicity.
\newline
Figure~\ref{fig:WINDSYN_timeseries} shows the temporal development of the generated wind power output, normalized by the plant capacity, of a single wind power plant for a three-week window as an example.
Due to the characteristic properties of the power curve, there is no feed-in power at very low wind speeds (ca. < 1-2 m/s).
If high wind speeds occur (ca. > 12 m/s), the generation is limited by the nominal power of the wind turbine.
These properties form the characteristic shape of such wind power time series.
\newline
At the same time, the yellow curve shows the forecasted wind speed.
Here we can observe that wind speed and power are correlated since higher predicted wind speeds also tend to have higher power values.
This correlation becomes even more evident in figure~\ref{fig:WINDSYN_scatter}, where the generated power, the synthetic measurement, is plotted above the predicted wind speed, given by the \ac{nwp}.
The typical S-shaped course of the power curve appears here.
At the same time, however, there is also a wide spread.
The spread, as in the case of \ac{pv}, is due to the forecast errors of the weather resource.
Analogously to \ac{pv}, this uncertainty is close to the real world and thus also shows up in actual wind power forecast data.





\section{Experimental Results}\label{sec_experimental_results}
The following section conducts an experiment that serves as the baseline for future research.
Moreover, the experiment allows comparing forecast errors of the proposed synthetic dataset with real-world forecast errors, showing our dataset's applicability for future research in renewable power forecasts.
We evaluate physical forecasting techniques on the introduced datasets within the following experiment.
Physical approaches are essential as they are typically utilized when only limited data is available for training.
At the same time, there is an increasing interest in providing high-quality forecasts through~\ac{ml} models even with limited data.
Therefore, we compare the results of physical models with the \ac{gbrt}.
We choose the \ac{gbrt} here as it generalizes well with a limited amount of data and is known to mitigate the effects of overfitting.

To assess the forecast quality of the~\ac{gbrt} concerning reduced data, as it is common with~\ac{tl}, we limit the training data to the last $7$,$14$,$30$,$60$ or $90$ days of training data of a season.
Respectively, we repeat this experiment for each season's different number of training data to ensure that results account for seasonal patterns.
To increase the number of training data within this scenario, we linearly interpolate the data to have a $15$-minute resolution.
Additionally, we make one experiment where we consider the complete training data, indicated through $365$, and compare those forecast errors with an example from real-world datasets.
We consider time steps with the test flag as test data and the remaining data as training data.

Most often the forecast error, in renewable power forecast, is determined by the~\ac{nrmse} given by
\begin{equation}\label{eq_nrmse}
    \text{nRMSE} = \sqrt{\frac{1}{N}\sum_{i=0}^{i=N}{(y_i-\hat{y}_i)^2}},
\end{equation}
where $y_i$ is the $i$-th normalized response from the target and $\hat{y}_i$ is the prediction from a source model on the target.
Note that in the context of renewable power forecasts, we normalize the response $y_i$ by the nominal power to assure comparability of the error for different parks.
To express an improvement of one model over another model we utilize the skill given by
\begin{equation}
    \text{Skill} = \text{nRMSE}_{gbrt}-\text{nRMSE}_{phy},
\end{equation}
where $\text{nRMSE}_{gbrt}$ is the \ac{nrmse} of the~\ac{gbrt} model and $\text{nRMSE}_{phy}$ of the physical model, respectively.
Consequently, values below zero indicate an improvement of the~\ac{gbrt} over a physical forecasting technique.

We optimize the~\ac{gbrt} for each park, season, and number of training data through a grid search by three-fold cross-validation based on the available training data.
Thereby, the learning rate is evaluated at $[\SI{e-6}, \num[round-precision=1]{3.1e-6}, \ldots,\num[round-precision=1]{3.1e-1}, 1]$.
The number of estimators is $300$ and the maximum depth is one of $[2,4,6,8]$.
Other values are the default ones of scikit-learn~\footnote{\url{https://scikit-learn.org/}, Version 0.24, accessed 2022-02-28}.

We utilize the same parametrization for the physical models for creating the power measurements.
The only difference is that we now use the direct radiation and wind speed, respectively, from the~\ac{nwp} as input.
The physical model for the wind dataset is indicated through \textit{Enercon} in the following and \textit{PV Physical} for the~\ac{pv} dataset.
Additionaly, for the wind dataset, we compare the~\ac{gbrt} results with empirical \textit{McLean} power curves~\cite{Mclean2008}.
This power curve is well known and considers typical influences by different terrains.

\begin{figure}[tb]
	\centering
	\includegraphics[width=\textwidth]{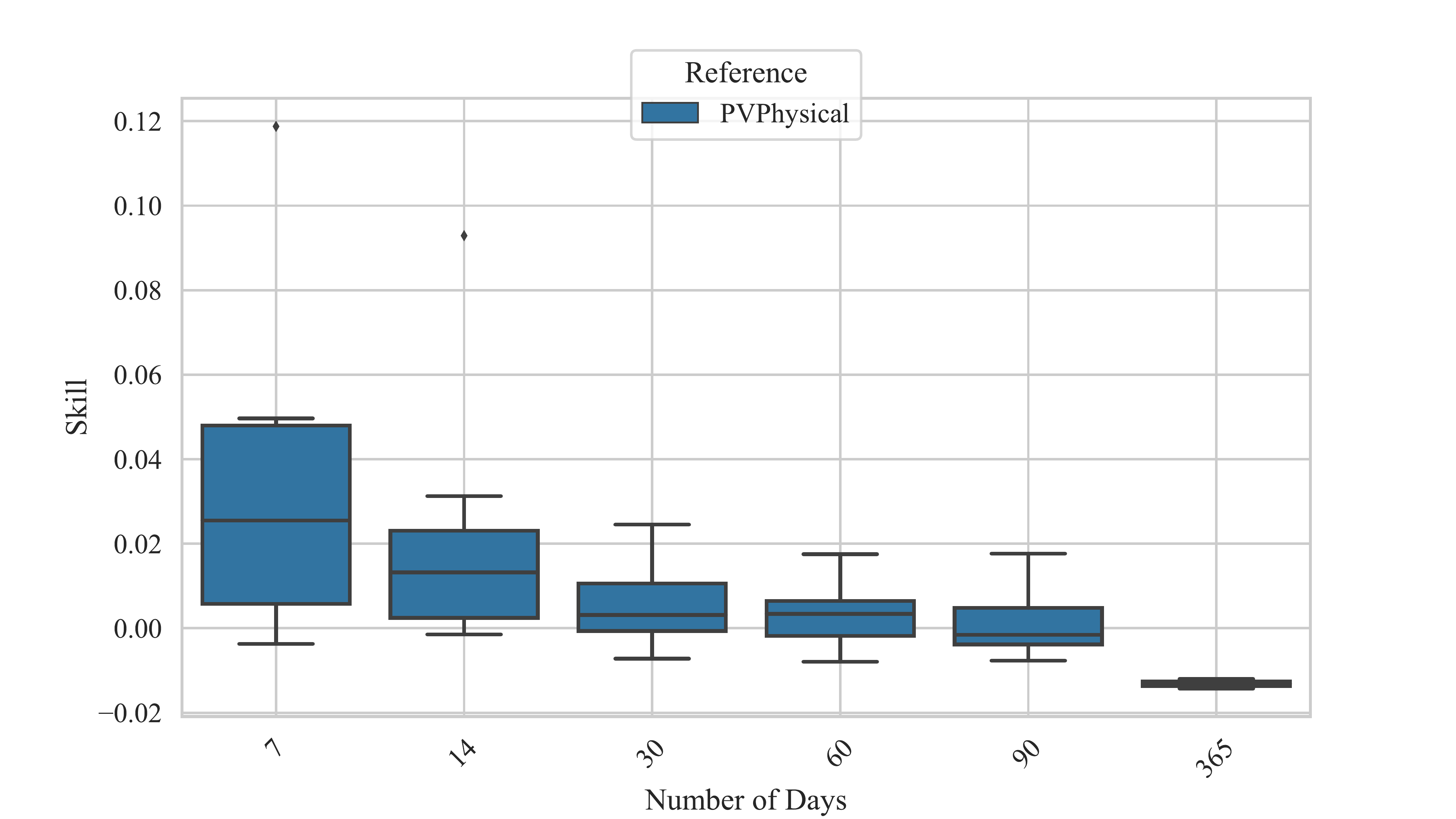}
	\caption{Skill between the GBRT model and the physical baseline of the dataset for the PV dataset. Values below zero indicate an improvement of the GBRT over the physical reference.}\label{pvsyn_physical_skill_phyiscal}
\end{figure}

PV day-ahead power forecasting problems often have an~\ac{nrmse} between $0.06$ and $0.12$~\cite{Schreiber2020Prophesy}.
The mean~\ac{nrmse} values of the PV baseline is at $0.085$ on the proposed synthetic dataset, see Tbl.~\ref{tbl_results_pv}.
Forecast errors of the~\ac{gbrt} model are between $0.12$ and $0.072$ depending on the amount of available training data.
With an increasing amount of training data, the forecast error reduces for this model.
The mean forecast error of the~\ac{gbrt} is only better with the complete training data.
Fig.~\ref{pvsyn_physical_skill_phyiscal} shows the evaluation concerning the skill for the PV dataset.
Up to $14$ days of training data, the~\ac{gbrt} model has substantial positive outliers of the skill.
With additional training data, this effect reduces. 
Beginning from $60$ days of training data, the median skill is close to zero, indicating the~\ac{gbrt} is as good as the physical baseline in most cases.
With $90$ days of training data and the complete dataset, the~\ac{gbrt} has improvements over the physical baseline.

\begin{table}[tb]
    \centering
    \caption{Mean nRMSE values of the~\ac{gbrt} across all available parks for different amounts of training data in comparison to physical models for the PV dataset. Best values for the amount of training data are highlighted in bold.}\label{tbl_results_pv}
    \begin{tabular}{lrr}
    \toprule
    {} &   GBRT &  PV Physical \\
    Number of Days &        &             \\
    \midrule
    7               &  0.120 &       \textbf{0.085} \\
    14              &  0.107 &       \textbf{0.085} \\
    30              &  0.092 &       \textbf{0.085} \\
    60              &  0.089 &       \textbf{0.085} \\
    90              &  0.086 &       \textbf{0.085} \\
    365             &  \textbf{0.072} &       0.085 \\
    \bottomrule
    \end{tabular}
\end{table}

\begin{figure}[tb]
	\centering
	\includegraphics[width=\textwidth]{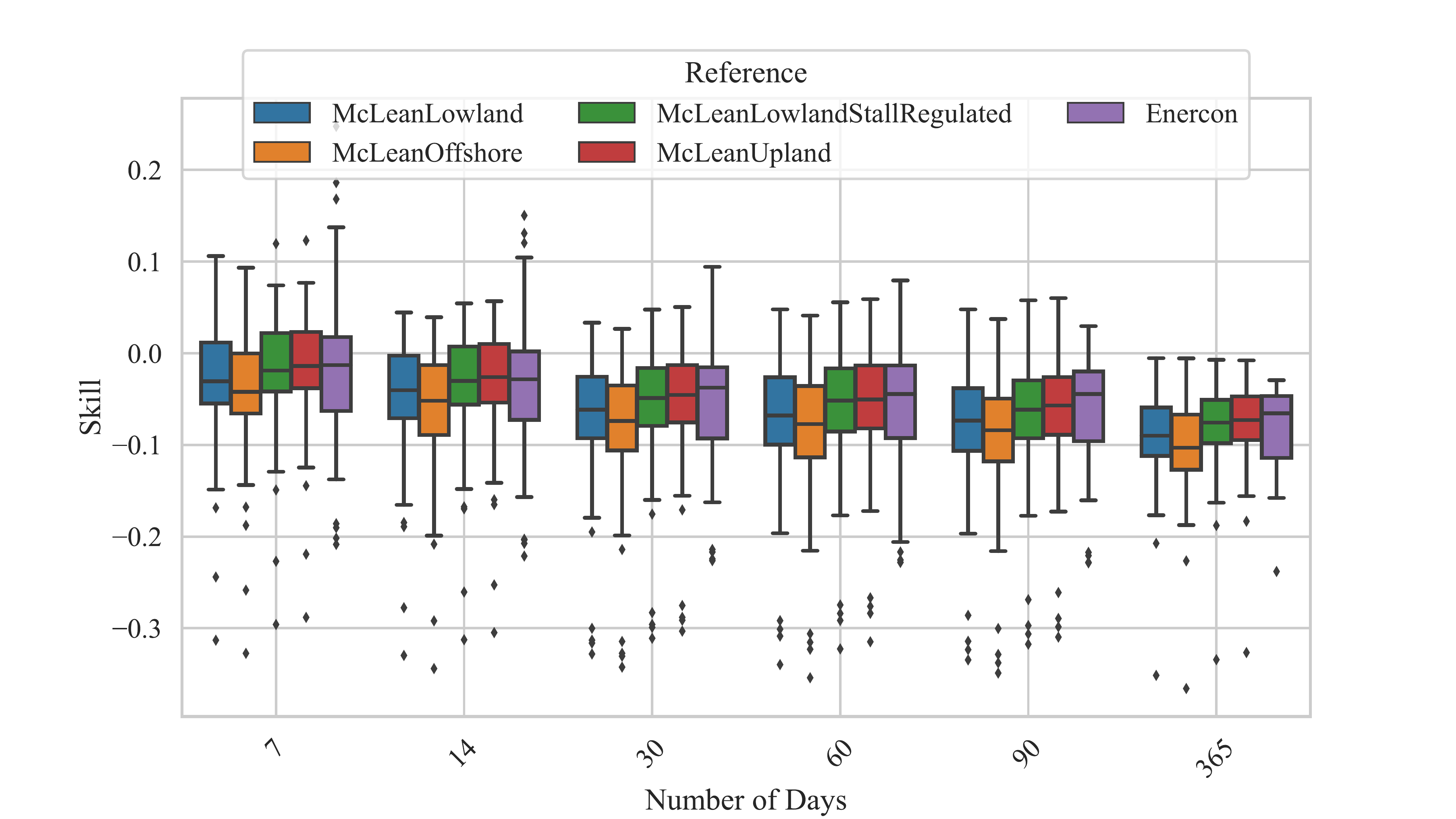}
	\caption{Skill between the GBRT model and the physical baseline (Enercon) and different McLean power curves for the wind dataset. Values below zero indicate an improvement of the GBRT over those references.}\label{windsyn_physical_skill_phyiscal}
\end{figure}

Tbl.~\ref{tbl_results_wind} summarizes the mean~\ac{nrmse} for the wind dataset.
The Enercon baseline has a mean~\ac{nrmse} of $0.21$.
Wind day-ahead power forecasting problems typically have an~\ac{nrmse} between $0.1$ and $0.2$~\cite{Schreiber2020Prophesy}.
Depending on the utilized empirical McLean power curve, forecast values of this approach range between $0.212$ and $0.228$.
The~\ac{gbrt} forecasting technique has the best mean~\ac{nrmse} for all numbers of available training data.
With seven days of training data, the~\ac{gbrt} has an~\ac{nrmse} of $0.196$, with $90$ days of training data a~\ac{nrmse} of $0.147$, and with the complete data, an error of $0.125$.
The improvements of the~\ac{gbrt} over the other models are also given through the skill in Fig.~\ref{windsyn_physical_skill_phyiscal}.
In contrast to results from the~\ac{pv} dataset, the median skill is below zero already with seven days of training data.
However, with less than $30$ days of training data, substantial outliers are present compared to the Enercon baseline.
These are not present with more than $30$ days of training data.

All in all, we can observe that the forecast error of the physical baseline and, in particular, errors from the~\ac{ml} technique are comparable to those given by real-world power measurements.
For the~\ac{pv} dataset, we see that the physical baseline is robust, especially in the case of limited data.
At the same time, these excellent results also relate to the fact that the exact physical characteristics are given, which is most often not the case in real-world forecasting problems.
In the case of the proposed wind dataset, the~\ac{gbrt} provides a strong baseline for future research.
In a comparison of those two datasets, the physical approach for the~\ac{pv} dataset considers more physical characteristics in comparison to the Enercon baseline.
At the same time, the~\ac{pv} power forecasting problem can be considered more linear in contrast to wind power forecasts, which makes it easier to model all dependencies within a physical baseline.
Interestingly, errors of the \textit{McLean-Upland} empirical power curve are close to that of the Enercon baseline.
This observation lets us conclude that it is a suitable baseline for limited data when a wind turbine's physical characteristics are not given.


\begin{table}[tb]
    \centering
    \caption{Mean nRMSE values of the~\ac{gbrt} across all available parks for different amounts of training data in comparison to the physical model and different McLean power curves for the wind dataset. Best values for the amount of training data are highlighted in bold.}\label{tbl_results_wind}
    \begin{tabular}{lrrrrrr}
    \toprule
    {} &   GBRT &  Enercon &  McLean- &  McLean- &  McLean- &  McLean- \\
    {} &    &   &  Lowland &  Lowland-Regulated &  Offshore &  Upland \\
    Number of Days &        &          &                &                              &                 &               \\
    \midrule
    7               &  \textbf{0.196} &    0.210 &          0.228 &                        0.216 &           0.239 &         0.212 \\
    14              &  \textbf{0.178} &    0.210 &          0.228 &                        0.216 &           0.239 &         0.212 \\
    30              &  \textbf{0.156} &    0.210 &          0.228 &                        0.216 &           0.239 &         0.212 \\
    60              &  \textbf{0.154} &    0.210 &          0.228 &                        0.216 &           0.239 &         0.212 \\
    90              &  \textbf{0.147} &    0.210 &          0.228 &                        0.216 &           0.239 &         0.212 \\
    365             &  \textbf{0.125} &    0.210 &          0.228 &                        0.216 &           0.239 &         0.212 \\
    \bottomrule
    \end{tabular}
\end{table}

\section{Conclusion and Future Work}\label{sec_conclusion_future_work}
We successfully created a realistic wind and photovoltaic power forecast dataset within the article.
Power measurements are generated based on real-world weather measurements, ideal for creating power time series through physical models.
At the same time, considering numerical weather prediction as input features for the \ac{ml} model, results show that forecast errors are in line with other real-world renewable power datasets.
By providing metadata for each plant, we provide the possibility for transfer learning, multi-task learning, and zero-shot learning for future research by considering those physical characteristics.

\textbf{Acknowledgments}
This work has been partially carried out within the project TRANSFER (01IS20020B) funded by BMBF (German Federal Ministry of Education and Research) and the research project gridcast (Fkz. 0350004A) funded by the BMBK (German Federal Ministry for Economic Affairs and Climate Action). Furthermore, we would like to express our special thanks to the German Weather Service (DWD) and the Fraunhofer Institute for Energy Economics and Energy System Technology (IEE) for providing the weather data.

\bibliographystyle{unsrt}






\end{document}